\documentclass{bmvc2k}

\usepackage{graphicx}

\usepackage{tikz}
\usepackage{comment}
\usepackage{amsmath,amssymb} 
\usepackage{color}
\usepackage{multicol}
\usepackage{multirow}
\usepackage{booktabs}
\usepackage{commath}
\usepackage{makecell}
\usepackage{capt-of}
\usepackage{etoolbox,refcount}
\usepackage{xspace}

\DeclareMathOperator{\sgn}{sgn}

\usepackage[colorlinks]{hyperref}

\usepackage[capitalize]{cleveref}
\crefname{section}{Sec.}{Secs.}
\Crefname{section}{Section}{Sections}
\Crefname{table}{Table}{Tables}
\crefname{table}{Tab.}{Tabs.}

\makeatletter
\DeclareRobustCommand\onedot{\futurelet\@let@token\@onedot}
\def\@onedot{\ifx\@let@token.\else.\null\fi\xspace}

\def\eg{\emph{e.g}\onedot} 
\def\ie{\emph{i.e}\onedot}

\def\wrt{w.r.t\onedot} 
 
\def\etal{\emph{et al}\onedot}
\makeatother

\title{Self-Improving SLAM in Dynamic Environments: Learning When to Mask}

\addauthor{Adrian Bojko}{adrian.bojko@cea.fr}{1}
\addauthor{Romain Dupont}{romain.dupont@cea.fr}{1}
\addauthor{Mohamed Tamaazousti}{mohamed.tamaazousti@cea.fr}{1}
\addauthor{Hervé Le Borgne}{herve.le-borgne@cea.fr}{1}

\addinstitution{
 Université Paris-Saclay, CEA, List \\
 F-91120, \\
 Palaiseau, France\\
}

\runninghead{Bojko \etal}{Self-Improving SLAM in Dynamic Environments}

\begin{document}

\maketitle

\begin{abstract}
Visual SLAM -- Simultaneous Localization and Mapping -- in dynamic environments typically relies on identifying and masking image features on moving objects to prevent them from negatively affecting performance. Current approaches are suboptimal: they either fail to mask objects when needed or, on the contrary, mask objects needlessly. Thus, we propose a novel SLAM that learns when masking objects improves its performance in dynamic scenarios. Given a method to segment objects and a SLAM, we give the latter the ability of Temporal Masking, \ie, to infer when certain classes of objects should be masked to maximize any given SLAM metric. We do not make any priors on motion: our method learns to mask moving objects by itself. To prevent high annotations costs, we created an automatic annotation method for self-supervised training. We constructed a new dataset, named ConsInv, which includes challenging real-world dynamic sequences respectively indoors and outdoors. Our method reaches the state of the art on the TUM RGB-D dataset and outperforms it on KITTI and ConsInv datasets.

\end{abstract}

\vspace*{-4mm}
\section{Introduction}
\label{sec:introduction}

\begin{figure*}[ht]
    \centering
    \includegraphics[width=1.0\linewidth]{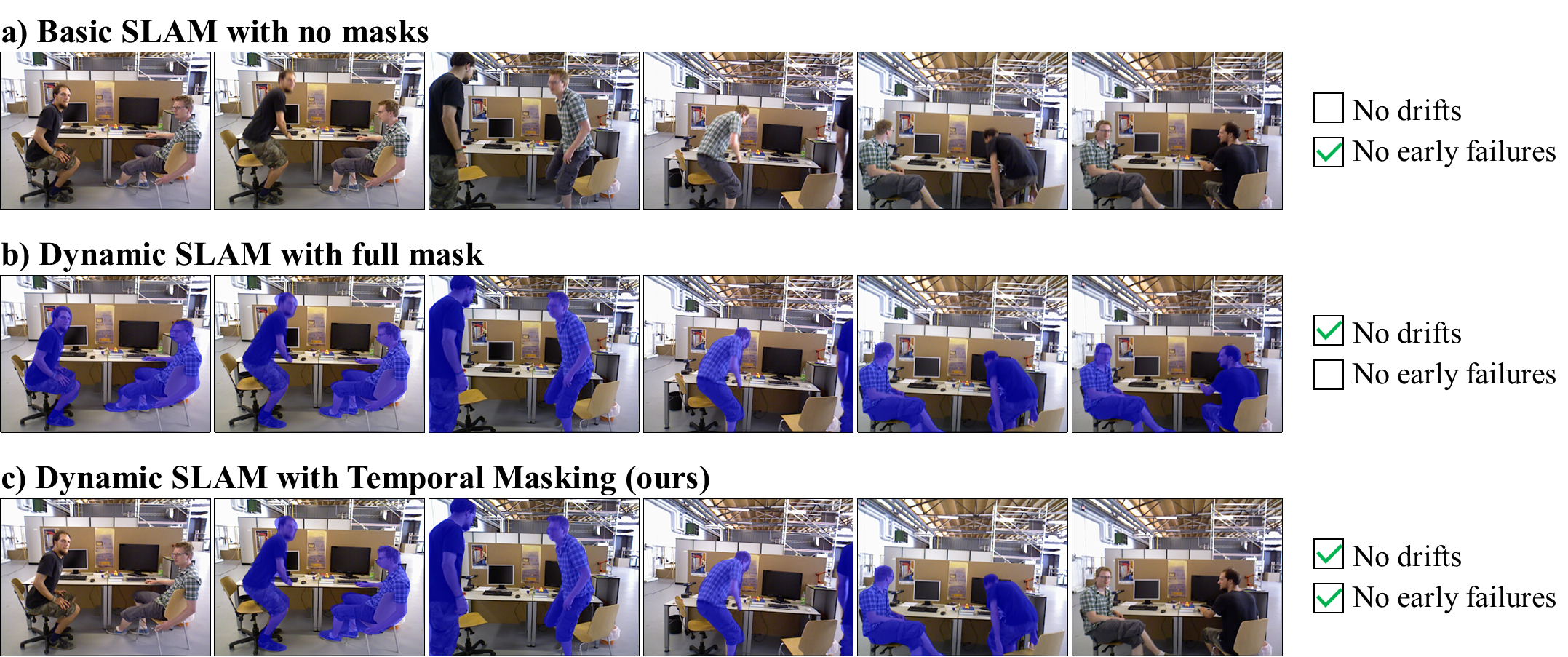} 
    \caption{Illustration of our results on the TUM RGB-D dataset. a) A Basic SLAM does not mask dynamic objects and consequently drifts. b) A Dynamic SLAM masks all supposedly dynamic objects, even when they are static, and may consequently fail if there are not enough unmasked features. c) We add Temporal Masking to Dynamic SLAM: we mask dynamic objects when appropriate, \ie, when we predict that masking them improves SLAM performance. Our model learns by itself that masking objects in motion is beneficial for the SLAM, while other approaches must make this assumption.}
    \label{fig:failures}
\end{figure*}

Visual Dynamic SLAM algorithms are camera-based Simultaneous Localization and Mapping algorithms that filter the visual features, based on direct motion detection \cite{lenz_sparse_2011,cheng_accurate_2018} or a combination of semantic segmentation and motion detection \cite{bescos_dynaslam:_2018,schorghuber_slamantic_2019,xiao_dynamic-slam:_2019,ballester_dot_2020,xiao_dynamic-slam:_2019,zhang_front-end_2021}.
However, when the motion of a dynamic object is dominant in the image, these approaches may fail due to motion consensus inversion~\cite{bojko_learning_2020}. Moreover, approaches that rely on semantic segmentation also tend to mask objects even if they are not moving (\eg, parked vehicles) -- this excessive masking causes failures due to lack of features. To address these issues, \cite{schorghuber_slamantic_2019} introduces a ``dynamic factor'' for each 3D point to identify those that should be used for pose estimation. It nevertheless fails under motion consensus inversion, as the instantaneous motion of an object is too difficult to detect geometrically, or when an object moves after being considered static. \cite{ballester_dot_2020} computes a camera motion using features that are static according to semantic segmentation but fails if too few features are located on static objects.

We propose a novel SLAM algorithm that learns when to mask object classes based on its own performance in dynamic environments, in a self-supervised fashion. Unlike existing approaches, our decision process is learned. We infer masking decisions that maximize SLAM performance \emph{separately} from the SLAM itself: we use only past frames to compute which classes to mask in the current image while not using any SLAM-dependent data -- keypoints, reconstructed map, camera pose, etc. This results in an improved robustness in dynamic scenarios, especially in challenging cases. We do not assume that certain classes (\eg, cars) or moving objects have to be masked as our models learn these facts by themselves. 

This paper shows the feasibility and interest of this new masking decision paradigm that we name \emph{Temporal Masking}. Our technical contribution is an automatic annotation method for self-supervised training and the corresponding neural architecture, which relies on recurrent neural networks to take into account possible dependencies with past frames. Given a SLAM algorithm and a method to generate semantic masks, we train a model that decides for every supported semantic class if objects of this class should be masked in the current frame, without any underlying geometrical considerations nor assumptions on what object classes are dynamic. Our approach is agnostic with respect to the choice of SLAM algorithm and semantic segmentation method. Since most effective learning-based approaches are supervised, we prefer such an approach although it requires annotations. To annotate a sequence, we must make a binary decision \texttt{mask/no\_mask} for every frame and every semantic class; we call such a set a temporal mask. Beyond the tediousness of the task, it is difficult to make such a choice \emph{a priori} with regard to the final SLAM performance, even for an expert, and it may be algorithm-dependent. Thus, we propose to create temporal masks automatically. As evaluating all possible combinations of masking decisions is computationally intractable, we uniformly sample a constrained space of all possible temporal masks, then evaluate and aggregate the samples according to their SLAM performance (\eg, accuracy). The complexity of the annotation process decreases from exponential to constant time \wrt the number of benchmarked labels per sequence. Our annotation method can use any performance metric, including the classical trajectory accuracy (ATE RMSE) and robustness (Tracking Rate). We propose a new metric that combines both and is consistent with a human user expectations.

We improved a SLAM algorithm to make it robust to the two points of failure of existing methods in difficult dynamic scenarios, by making it decide whether applying a given spatial mask is relevant or not with regard to the final SLAM performance. It is on par or beyond the state of the art on TUM RGB-D and KITTI datasets. It is also beyond it on a new dataset we built, ConsInv, which includes motion consensus inversions and the risk of failure due to excessive masking, rarely present in usual benchmarks. We also report the limits of our approach when the test context is different from the training one.

\vspace*{-4mm}
\section{Related Work}
Saputra \etal \cite{saputra_visual_2018} gives an overview of Dynamic SLAM: the general principle is to remove image features on dynamic objects so that the SLAM only uses static features. Current methods detect motion, semantically segment dynamic objects, or do both. Motion-based masking approaches filter features in dynamic regions \cite{lenz_sparse_2011,cheng_improving_2019,xu_mid-fusion_2019,scona_staticfusion:_2018,cheng_accurate_2018,sun_improving_2017,li_rgb-d_2017,sun_motion_2018}, typically with optical flow or clustering approaches. A flaw of such approaches is the underlying prior that most of the image corresponds to static objects. In case of motion consensus inversion, static objects are considered dynamic and vice-versa, making the SLAM map only dynamic objects, thus drifting when the objects move. Such inversions occur when most of the features are on dynamic objects, which may occur when they are too close to the camera. Semantic masking approaches \cite{kaneko_mask-slam:_2018} typically use networks as Mask R-CNN \cite{he_mask_2017} to filter features on objects of specific classes (\eg, cars). They are unlikely to cause SLAM drifts as semantic segmentation does not rely on motion consensus, unless there are unknown dynamic object classes. However, they often mask objects even if they are not moving (\eg, parked cars) and cause SLAM failure when there are not enough remaining features.

Combined approaches combine semantic and motion masking. \cite{bescos_dynaslam:_2018} combines multiview geometry and semantic segmentation to remove features on dynamic objects of known and unknown classes. However, it may remove too many features as it masks all detected objects and fail. Some combined approaches apply semantic masks only when they consider the masked object dynamic. \cite{schorghuber_slamantic_2019} classifies features as static/dynamic according to the semantic class of the object they are on and how stable the positions of the corresponding map points are. However, features on a dynamic object may be mistakenly considered static under motion consensus inversion, making the method fail. Moreover, each semantic class is manually assigned a probability of moving, while we learn when an object is likely to move. \cite{ballester_dot_2020} computes a camera motion using features that are static according to the semantic segmentation, selects the dynamic features on motionless objects using photometric error and inputs both static and selected dynamic features into the SLAM backend. However, if there are features only on objects of dynamic classes, it will be unable to compute the camera pose and fail. \cite{xiao_dynamic-slam:_2019} is similar to \cite{ballester_dot_2020} but uses reprojection errors instead of photometric errors. \cite{barnes_driven_2018,zhang_front-end_2021,bojko_learning_2020} are two-step approaches: they learn to segment dynamic objects at training time and do semantic masking at runtime, so they may suffer from failures due to excessive masking. 

Current Visual Dynamic SLAM algorithms are unable to handle both motion consensus inversions and failures due to excessive masking. A key cause is the dependency on instantaneous motion detection, unreliable in difficult scenarios. More generally, the dependence on ad hoc conditions limits generalization to challenging scenarios or unknown contexts. Our approach is different as we infer the effect of masking objects on SLAM performance with temporal masking. We consider the SLAM as a black box with respect to masking decisions; so Dynamic SLAM algorithms that use internal SLAM data for object masking \cite{ballester_dot_2020,bescos_dynaslam:_2018,schorghuber_slamantic_2019,xiao_dynamic-slam:_2019} are incompatible. 

We tackle difficult scenarios where existing methods \cite{mouragnon2006real, klein_parallel_2007, tamaazousti2011real, tamaazousti2011nonlinear, tamaazousti2016constrained} do not work well, including Visual-Inertial SLAM as \cite{orbslam3_campos_2021, coulin2021tightly}. IMUs cannot be used for SLAM or odometry in non-inertial frames of references: airplanes, ships, trains, submarines, or space. Filtering dynamic objects in LiDAR point clouds is not trivial without robust vision – moreover, they cannot be used everywhere (\eg, medical / military contexts). We propose a method that makes camera-only visual SLAM robust in difficult scenarios that are poorly handled today. Our aim is to prevent critical failures (i.e., accidents) without the convenience of IMUs or other sensors. Finally, a better and cleaner vision signal benefits multi-sensor SLAM fusion.

\vspace*{-4mm}
\section{Temporal Masking}
\vspace*{-1mm}
\subsection{System Overview}
\begin{figure*}[ht]
    \centering
    \includegraphics[width=1.0\linewidth]{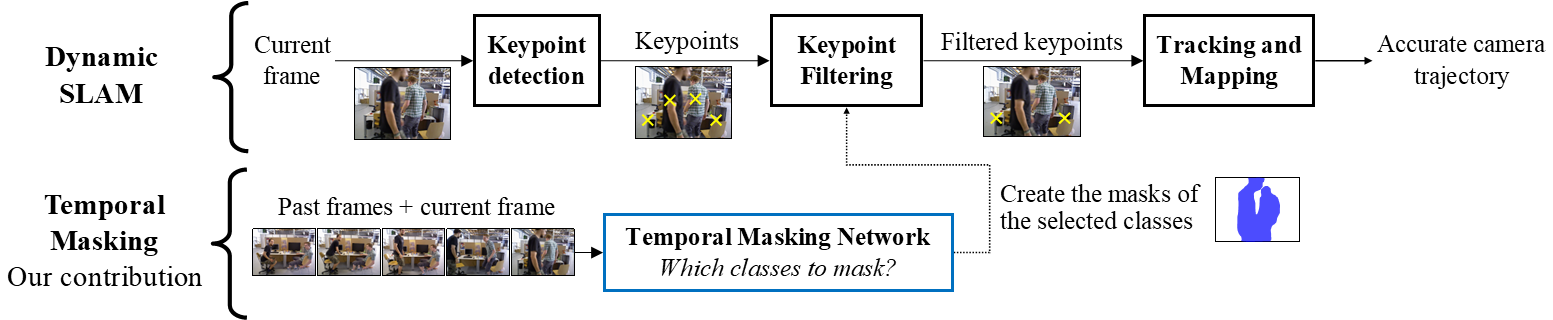}
    \vspace*{-4mm}
    \caption{Dynamic SLAM with Temporal Masking: to compute which semantic classes should be masked to maximize SLAM performance even in challenging dynamic scenarios. The resulting trajectory is as accurate as possible given the base SLAM and semantic segmentation network.}
    \label{fig:temporal_masking_pipeline}
\end{figure*}

Our approach can be applied to any Dynamic SLAM approach based on keypoint detection then filtering (Fig. \ref{fig:temporal_masking_pipeline}, top row). This last step usually consists in applying a spatial semantic mask to determine which keypoints to ignore when estimating the camera trajectory. Our approach consists in adding a Temporal Masking Network that decides whether the semantic mask should be applied or not to the current frame, with regard to the past sequence. This network is trained through self-supervision, as detailed in the next section, such that it memorizes for every class the circumstances when masking is beneficial for the SLAM. At runtime, it infers masking decisions for unknown sequences. If it infers that masking a certain class is beneficial to the SLAM, we filter keypoints on the corresponding masked area. We focus on feature-based SLAM as image masking is straightforward, but since our method is SLAM-agnostic, it can be applied to direct SLAM algorithms as \cite{engel_lsdslam_2014}. The pseudocode of our method is in the supplementary materials.

\begin{figure*}[ht]
    \centering
    \includegraphics[width=1.0\linewidth]{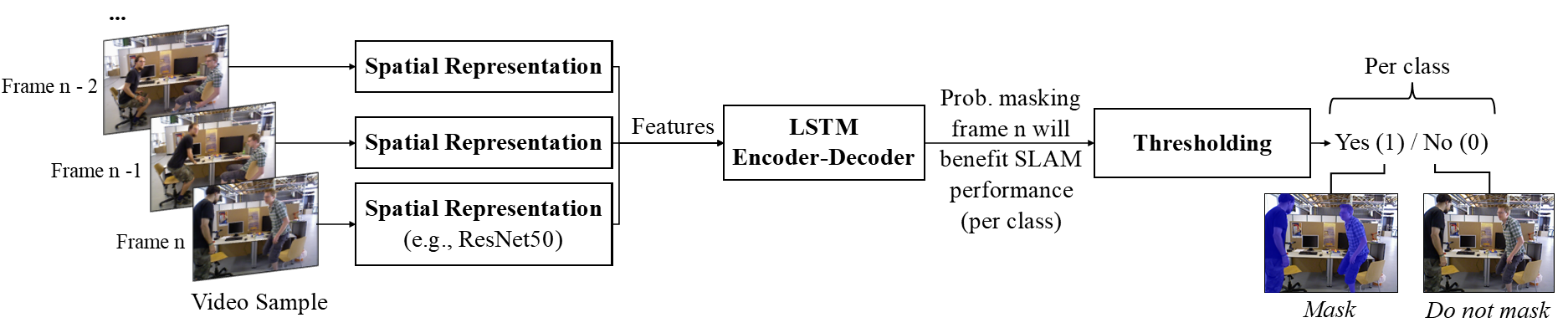}
    \vspace*{-4mm}
    \caption{Temporal Masking Network. It takes as input a video sample and outputs, for the last sampled frame, per-class binary masking decisions on which classes should be masked. Spatial Representation outputs features for every input frame.
    }
    \label{fig:temporal_neural_network}
\end{figure*}

The proposed architecture for the Temporal Masking Network is presented in \cref{fig:temporal_neural_network}. Our rationale is that relying on memory instead of geometry makes it possible to overcome the points of failure of current Dynamic SLAMs, unlike geometrical approaches that cannot understand the long-term context of a scene.
We designed the network for multi-label classification, in which every output corresponds to a semantic class we may need to mask. It is composed of 1) a spatial representation module that encodes input frames and 2) an LSTM Encoder-Decoder that makes memory-based decisions on which semantic classes to mask. The network takes as input a video sample, computes a spatial representation of every frame of the sample, inputs the computed features into the LSTM Encoder-Decoder, and finally applies a threshold to the result to obtain a per-class binary masking decision for the last frame of the sample. A possible way to compute features is to use a CNN encoder. We train only the LSTM Encoder-Decoder; we describe its architecture in supplementary materials.

\vspace*{-4mm}
\subsection{Temporal Annotation with Self-Supervision}
\label{sec:temporal_annotation_method}

Training a network for temporal masking requires annotating video sequences accordingly.
Let us consider a video sequence of length $l$ and a generator of semantic masks (\eg, Mask R-CNN) segmenting $p$ classes: a \emph{temporal mask} is a binary matrix of size $l \times p$ storing masking decisions (\ie, which classes to mask using the given generator) for every frame of the sequence. We first present the annotation method for a single class to mask. Note that iterating through the full temporal mask space is computationally intractable (up to $2^\text{seq. length}$ masks large), so we cannot exhaustively evaluate all temporal masks. Therefore, we compute temporal masks in three steps for every sequence: 1) Uniform random sampling of $q$ samples from a subset of all possible temporal masks. Sampling from a subset instead of the full space makes the problem computationally tractable. 2) Benchmarking of the sampled temporal masks using a unified metric. 
3) Performance-weighted aggregation of sampled temporal masks into a unique mask. Samples that perform well tend to mask objects more appropriately, so the result from the aggregation masks objects precisely when appropriate. Step 2 is straightforward: for every sample, we run the SLAM applying semantic masks according to the masking decisions in the sample and measure its performance with a unified metric. We detail steps 1) and 3) in the single-class case, then generalize to multiclass.

\begin{figure*}[ht]
    \centering
    \includegraphics[width=1.0\linewidth]{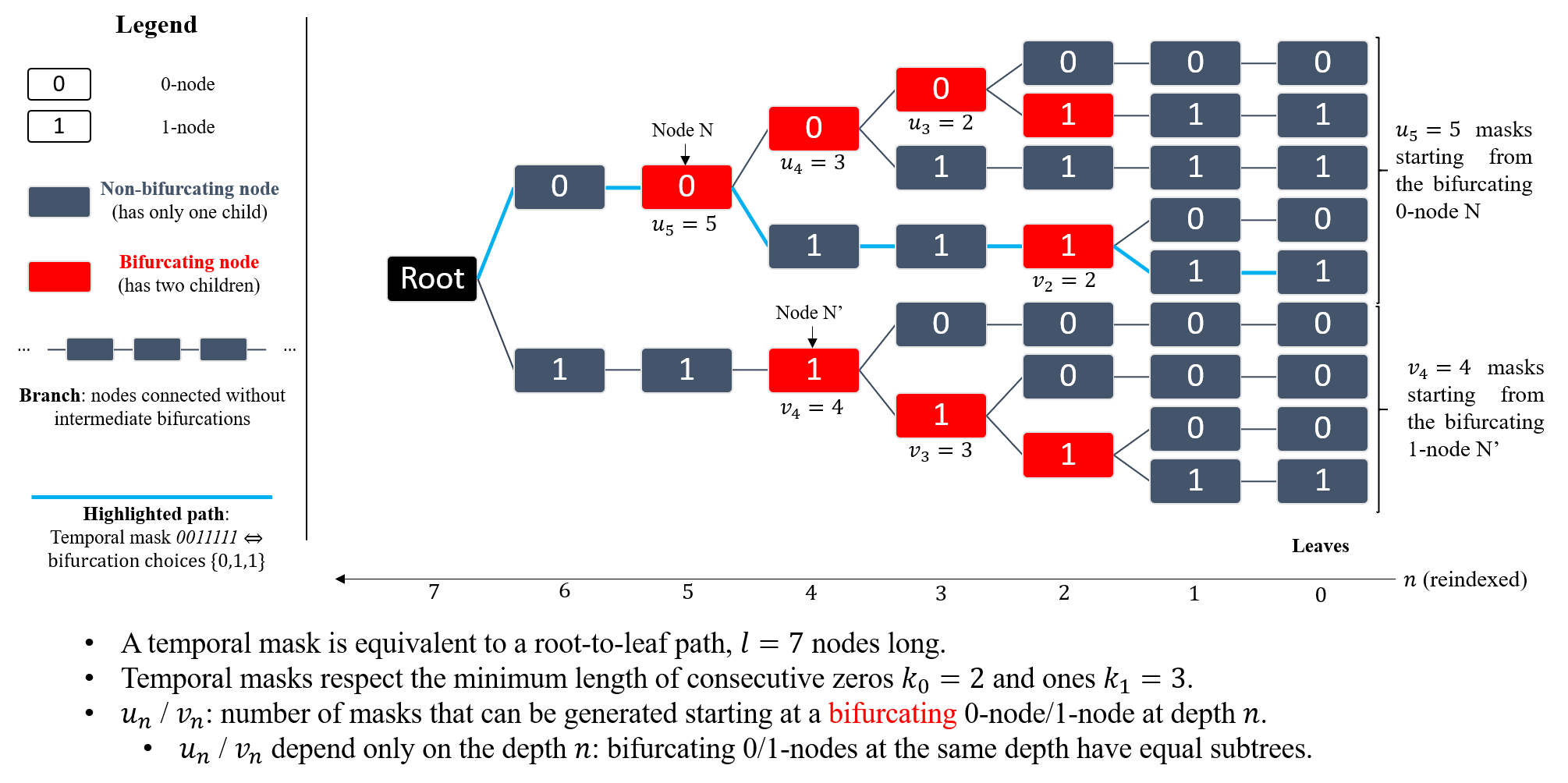} 
    \caption{Temporal mask space $E(l=7,k_0=2,k_1=3)$ as a masking binary tree. A temporal mask is equivalent to a root-to-leaf path.}
    \label{fig:temporal_mask_tree_short}
\end{figure*}

\textbf{Random Sampling of Temporal Masks Subspace.} We generate a set of basic temporal masks that we will benchmark to know their impact on SLAM performance. Let $l$ the sequence length, $k_0$ and $k_1$ the min. required length of resp. contiguous blocks of zeros and blocks of ones in a temporal mask (0 = \textbf{do not mask}, 1 = \textbf{mask}). For instance, if $l=7$, $k_0=2$ and $k_1=3$, then the masks \emph{1110000} and \emph{0011100} respect $k_0$ and $k_1$, but \emph{111\textcolor{red}{0}111} and \emph{00\textcolor{red}{11}000} do not. We uniformly sample masks from the space $E(l,k_0,k_1)$ of all temporal masks of length $l$ that respect the min. lengths $k_0$, $k_1$. The rationale is that the states of the objects in the scene do not change too quickly: by skipping high-frequency changes (low $k_0$, $k_1$), we focus on masks more suited to the scene. If $k_0=k_1=1$ the sampling is trivial. We explore $E$ with a binary tree where node value/depth $\Longleftrightarrow$ frame label/index (see Fig. \ref{fig:temporal_mask_tree_short}).

To keep the problem computationally tractable we cannot compute the whole tree of possible masks. The key is to use a closed-form expression of the number of possible paths (\ie, masks) $C(N)$ under any bifurcating node $N$. Starting at the root, whenever we reach a node that has two children, we randomly pick a child with the odds of each child proportional to the number of possible masks under it. To compute a closed-form solution for $C$, we define $u_n$ and $v_n$ as the number of masks under resp. bifurcating nodes of value 0 and 1 at depth $n$, reindexed from the end of the tree. Let $k:=k_0+k_1$. Then:
 \resizebox{.9\linewidth}{!}{
  \begin{minipage}{\linewidth}
    \begin{equation}
        \label{eq:linear_relations_short}
        \begin{cases}
            u_{n} = 2u_{n-1} - u_{n-2} + v_{n-k-1}\\
            v_{n} = 2v_{n-1} - v_{n-2} + v_{n-k-1}
        \end{cases}
    \end{equation}
  \end{minipage}
}

\eqref{eq:linear_relations_short} shows that $(u_n)$, $(v_n)$ are linear recurrence relations with constant coefficients and the same characteristic equation: $x^{k+1}-2x^k+x^{k-1}-1=0$. Thus, we can numerically compute a closed-form solution for $(u_n)$, $(v_n)$, and $C(N)$. Full proof and details in supp. mats.

\textbf{Sampled Temporal Mask Aggregation.} After measuring the performance (\ie, score) of all temporal masks previously sampled, we aggregate them. We sum the differences between all pairs of sampled masks (sampled masks are binary vectors made of 0/1, so their differences are vectors made of -1/0/1) weighted by their score difference. The idea is that a difference in performance is explained by the difference in masking decisions, so the score-weighted sum is a real vector where high values indicate frames that must be masked and low values frames that must not be masked~\cite{smeaton2005imvip}. Let $x,y$ be two sampled temporal masks and $s_x,s_y$ their respective scores. To consider only  significant score differences, let $\sigma_a$ be the absolute noise (below which score differences are meaningless) and $\sigma_r$ the relative noise (the score difference must be at least $\sigma_r$ times the first score of the pair). Then we compute the result vector $R$: \\ 
\resizebox{0.85\linewidth}{!}{
  \begin{minipage}{\linewidth}
    \begin{equation}
    \begin{aligned}
        R=\sum_{x,y\in\textnormal{Samples}}
            \begin{matrix}
                \max(0,\sgn(\abs{s_y-s_x}-\max(\sigma_r\abs{s_x},\sigma_a)))\\
                \times (s_y-s_x)(y-x)
            \end{matrix}
        \end{aligned}
    \end{equation}
  \end{minipage}
}

$R$ is a real-valued vector that we normalize in $[0,1]$ then binarize with thresholds in $[0,1]$, generating an arbitrary number of masks. We test them and select the best one score-wise. If there are equivalent masks within $\sigma_a$ and $\sigma_r$, we select the one that masks the most frames.

\textbf{Generalization to multiple classes.} Let a sequence of length $l$ and $p$ classes to mask. In the single-class case, we sample $q$ vectors of length $l$. In the multiclass case, we sample $pq$ vectors that we join in matrices of size $l \times p$. The benchmark step is unchanged. The aggregation step is the same up to the computation of $R$, which is now a real-valued matrix. In the single-class case, $R$ is a vector that we binarize by applying thresholds. In the multiclass case, every class (\ie, column) may have a different optimal threshold. Hence, for every class $i$: 1) We generate a matrix $R_i$ by zeroing out all columns other than column $i$. 2) As in the single-class case, we apply thresholds, generating an arbitrary number of temporal masks, evaluate them and select the best one $T_i$. 3) We select the best temporal mask score-wise. 3) We concatenate columns $1,\dots,p$ of resp. $T_1,\dots,T_p$, resulting in a temporal mask where all classes are appropriately masked.

\vspace*{-3mm}
\subsection{USM: Unified SLAM Metric}
\label{sec:usm}
ATE RMSE (Absolute Trajectory Error) and sometimes Tracking Rate (\% of tracked frames) are two SLAM metrics of interest for our method since they resp. reflect SLAM accuracy and robustness. 
As our method maximizes a single scalar value, we propose a metric to unify both ATE RMSE and Tracking Rate (TR), $\text{USM}_\lambda$:
\vspace*{-2mm}
\begin{equation}{\label{eq:usm_metric}}
    \text{USM}_\lambda(\text{ATE}, \text{TR}) = \text{TR} \times e^{-\lambda \times \text{ATE}}
\end{equation}

$\lambda$ balances ATE RMSE and Tracking Rate while  ensuring dimensional consistency. If TR=100\% and $\text{ATE} \ll \frac{1}{\lambda}$, then $\text{USM}_\lambda \sim 1-\lambda \times \text{ATE}$, which is consistent with the usual ATE RMSE metric. If the system fails early (low TR), the score is penalized proportionally. Thus, the general behavior of our metric corresponds to user expectations. 

A practical way to ensure $\text{ATE} \ll \frac{1}{\lambda}$ while balancing ATE RMSE/TR is to set $\lambda=\frac{0.1}{\text{Avg. dataset ATE RMSE}}$. Another benefit of the USM is to make comparisons between SLAM algorithms easier to interpret even in very difficult scenarios, \eg, comparing (ATE RMSE=1mm, TR=10\%) to (ATE RMSE=1cm, TR=100\%). Existing approaches that take failures into account like the AUC \cite{async_yang_2021} do not balance both metrics.

We developed the USM for automatic data annotation. It is, however, practical for comparing SLAM algorithms: while multi-dimensional comparison is difficult, USM is a scalar whose balance between ATE RMSE and Tracking Rate is adjusted with the parameter $\lambda$. Moreover, using ATE RMSE / Tracking Rate alone may be misleading: consider a full sequence run with 10cm accuracy and a run that crashes after 10\% of the sequence: the accuracy is likely better in the latter case as there is less time for error accumulation. However, the USM takes such failures into account. More explanations in supplementary materials.

\vspace*{-4mm}
\section{Experiments}
\label{sec:experiments}

\vspace*{-2mm}
\subsection{Experimental setup}
\label{sec:experimental_setup_single_class}

\textbf{Datasets.} We evaluate the 8 dynamic sequences of TUM RGB-D dataset \cite{sturm_benchmark_2012} in RGB-D, a reference benchmark for Dynamic SLAM, the first 11 sequences of KITTI odometry dataset \cite{geiger_kitti_2012} in stereo, the ConsInv-Indoors dataset in monocular and the ConsInv-Outdoors dataset in stereo. For TUM RGB-D and KITTI, given the small number of sequences, we run our tests in a leave-one-out approach: we run the annotation/train pipeline on all sequences but one and infer/test temporal masks for the remaining one in a round-robin fashion. We propose the ConsInv dataset to evaluate Dynamic SLAM methods in very challenging conditions, including consensus inversions, risk of excessive masking and false starts. Difficulty levels and failure conditions are controlled (details and illustrations in supp. mats).

\textbf{Spatial representation.} We use ResNet-50 \cite{he_resnet_2016}
(TensorFlow 2, pretrained on ImageNet) to compute frame features. For every frame, we collect the output of each of the 17 convolutional blocks of ResNet-50. For each block, we apply a 2D average pooling on its output (into size $(4,4,.)$), flatten and apply Principal Component Analysis, computed per block on the training sequences. It results in a feature vector of length about 100 to 1000 per convolutional block, which we concatenate, resulting in a vector of size 15k to 20k. 

\textbf{LSTM Encoder-Decoder and Training.} Given a frame index $k$, we generate video samples by randomly sampling $49$ frames from frames $0$ to $k-1$ and adding frame $k$. We sample frames from the whole past to avoid depending on instantaneous motion detection (see \cref{sec:introduction}). We mask the first 49 frames as we cannot sample them without repetition. We train the LSTM Encoder-Decoder on train+val with early stopping. Training lasted 40 to 60 epochs in our experiments. We augment training sequences with horizontal flipping and process video samples in a random order. 

\textbf{SLAM and Semantic Masking.} We use ORB-SLAM 2 \cite{mur-artal_orb-slam2:_2017} (3000 features). As ORB-SLAM 2 is non-deterministic, we run it five times on KITTI dataset and ten times on other datasets for every sequence/temporal mask to evaluate and report the median score. We generate semantic masks for KITTI/ConsInv-Outdoors datasets using Mask R-CNN \cite{he_mask_2017}, trained on the COCO dataset \cite{coco_2014}, and for ConsInv-Indoors/TUM RGB-D datasets we used the DeepLabv3+ \cite{chen_encoder-decoder_2018} models from \cite{bojko_learning_2020}. 
We segment people/vehicles (7 classes) for KITTI, people and cars for ConsInv-Outdoors (2 classes), people for TUM RGB-D (1 class), and drom / lambo / dragon for ConsInv-Indoors (1 class\footnote{The DeepLabv3+ model we trained segments dromedary / lambo / dragon in a single \emph{dynamic} class.}). 

\textbf{Dataset Annotation and Metrics.} For automatic annotations (\ie, self-supervision), we generate $200$ temporal masks for each sequence with min. block sizes $k_0=k_1=25$ and use the USM (\cref{sec:usm}) to benchmark/aggregate temporal masks. As the ATE RMSE of sequences in the datasets TUM RGB-D/ConsInv and KITTI is resp $\approx$1cm and $\approx$1m with manual expert labels, we set $\lambda=10m^{-1}$ for TUM RGB-D/ConsInv and $\lambda=0.1m^{-1}$ for KITTI, which makes $USM\approx TR (1-\lambda \text{ATE})$ and comparisons between SLAM algorithms match user expectations. We compare annotation methods in supp. materials (full/weak supervision), where we show that automatic annotations are easier to learn that human ones.

\subsection{Comparison with the State of the Art}

\begin{table*}[ht]
    \centering
    \resizebox{\linewidth}{!}{
    \begin{tabular}{|c|c|c|cc|cccc|c|}
    \hline 
        \multirow{3}{*}{\textbf{Mode}} & \multirow{3}{*}{\textbf{Dataset}} & \multirow{3}{*}{\textbf{Metric}} & \multicolumn{2}{|c|}{\textbf{Baselines}} & \multicolumn{4}{|c|}{\textbf{State of the Art}} & \multirow{3}{*}{\textbf{Ours}} \\ 
        & & & No masks & Full masks & \makecell{Optical\\flow \cite{cheng_improving_2019}} & DynaSLAM \cite{bescos_dynaslam:_2018} & Slamantic \cite{schorghuber_slamantic_2019} & StaticFusion \cite{scona_staticfusion:_2018} & \\ \hline
        
        \multirow{3}{*}{RGB-D} & \multirow{3}{*}{TUM RGB-D} & ATE RMSE (m) ↓ & 0.105 & \textbf{0.019} & - & \textbf{0.019} & 0.028 & 0.099 & \textbf{0.019}\\ 
        & & Tracking Rate ↑ & \textbf{96\%} & \textbf{96\%} & - & 69\% & \textbf{96\%} & \textbf{96\%} & \textbf{96\%} \\ 
        & & USM ↑ & 0.55 & \textbf{0.80} & - & 0.57 & 0.76 & 0.54 & \textbf{0.80} \\ \hline
        
        \multirow{3}{*}{Stereo} & \multirow{3}{*}{KITTI} & ATE RMSE (m) ↓ & 2.59 & 2.67 & - & 2.74 & 2.70 & - & \textbf{2.51}  \\ 
        & & Tracking Rate ↑ &\textbf{100\%} & \textbf{100\%} & - & \textbf{100\%} & \textbf{100\%} & - & \textbf{100\%} \\
        & & USM ↑ & 0.80 & 0.80 & - & 0.79 & 0.80 & - & \textbf{0.81} \\ \hline

        \multirow{3}{*}{Stereo} & \multirow{3}{*}{ConsInv-Outdoors} & ATE RMSE$^*$ ↓ & 0.084 & \underline{0.019} & - & 0.025 & 0.032 & - & 0.024 \\
        & & Tracking Rate$^*$ ↑ & \underline{100\%}  & 75\% & - & 74\% & 85\% & - & 88\% \\
        & & USM ↑ & 0.61  & 0.81 & - & 0.80 & 0.82 & - & \textbf{0.88} \\
        \hline
        
        \multirow{3}{*}{Mono} & \multirow{3}{*}{ConsInv-Indoors-Dynamic} & ATE RMSE$^*$ ↓ & 0.074 & \underline{0.003} & 0.050 & 0.010 & 0.032 & - & 0.014 \\
        & & Tracking Rate$^*$ ↑ & \underline{94\%} & 74\% & 74\% & 70\% & 84\% & - & 84\% \\
        & & USM ↑ & 0.57 & 0.71 & 0.49 & 0.63 & 0.68 & - & \textbf{0.75} \\ \hline
        
        \multirow{3}{*}{Mono} & \multirow{3}{*}{ConsInv-Extra-MeetingRoom  (\emph{domain shift})} & ATE RMSE$^*$ ↓ & 0.170 & 0.012 & 0.077 & \underline{0.011} & 0.077 & - & 0.012 \\
        & & Tracking Rate$^*$ ↑ & \underline{96\%} & 73\% & 62\% & 66\% & 86\% & - & 76\% \\
        & & USM ↑ & 0.33 & 0.65 & 0.34 & 0.60 & 0.54 & - & \textbf{0.67} \\ \hline
        
        \multirow{3}{*}{Mono} & \multirow{3}{*}{ConsInv-Extra-LivingRoom  (\emph{domain shift})} & ATE RMSE$^*$ ↓ & 0.091 & \underline{0.012} & \underline{0.012} & 0.020 & 0.016 & - & 0.013 \\ 
        & & Tracking Rate$^*$ ↑ & \underline{96\%} & 82\% & 62\% & 71\% & 84\% & - & 85\% \\
        & & USM ↑ & 0.51 & 0.73 & 0.55 & 0.60 & 0.69 & - & \textbf{0.74} \\ \hline
        
        Mono & ConsInv-Indoors-Static & \makecell{Prevented\\false starts}↑ & 56\% & \textbf{100\%} & 67\% & \textbf{100\%} & 78\% & - & \textbf{100\%} \\ \hline
    \end{tabular}
    }
    \caption[caption]{Comparison with the state of the art on various datasets in their preferred mode. Best in \textbf{bold}. Best ATE/TR results that are contradictory are \underline{underlined} (USM combines ATE RMSE/TR to solve this contradiction). '-' indicates that the mode is not supported by the SLAM algorithm. Scores are medians over the dataset (every sequence is evaluated at least 5 times). For ATE RMSE, lower is better (↓). For Tracking Rate / USM / Prevented false starts, higher is better (↑). We use our model trained on ConsInv-Indoors when evaluating our method on ConsInv-Extra to measure how it performs in new domains. *Difficult sequences in ConsInv dataset makes the lone interpretations of ATE RMSE/TR misleading (\cref{sec:usm}). }
    \label{tab:sota_all}
\end{table*}

\label{sec:comparison_sota_single_class}
We compare our method to DynaSLAM\footnote{DynaSLAM crashes on TUM RGB-D seq. fr3\_w\_xyz. We compute the avg. on the 7 others in this case.} \cite{bescos_dynaslam:_2018}, Slamantic \cite{schorghuber_slamantic_2019}, Lucas-Kanade optical flow \cite{cheng_improving_2019} (only in monocular), all based on ORB-SLAM 2. We also evaluate StaticFusion \cite{scona_staticfusion:_2018}, that supports only RGB-D. \emph{No masks} refers to the original ORB-SLAM2, \emph{Full masks} refers to ORB-SLAM 2 with semantic masks always applied. All methods use the same semantic masks. 
Note that median USM is not computed from median ATE RMSE and Tracking Rate. 

\textbf{TUM RGB-D dataset.} \cref{tab:sota_all} shows that the major obstacles for the SLAM for this dataset are motion consensus inversions, and our self-supervised model learned to mask objects most of the time to prevent them. 
All methods performed well, close or equal to the manual annotations, except DynaSLAM that performed barely above the \emph{No masks} baseline and StaticFusion below it. DynaSLAM removes even more features than the Full masks approach, which delays SLAM initialization and causes temporary tracking loss. The Tracking Rate of DynaSLAM is in fact exceedingly low despite a good accuracy (\ie, low ATE RMSE), which proves the importance of measuring the Tracking Rate and the relevance of our metric, the USM. StaticFusion fails in dynamic sequences with fast rotations, as it is unable to filter dynamic objects quickly enough. 

\textbf{KITTI dataset.} \cref{tab:sota_all} shows that our method slightly outperforms others as it makes slightly better masking decisions, and that the KITTI dataset is unsuitable for Dynamic SLAM testing since \emph{no matter when objects are masked}, performance hardly changes. In particular, the Tracking Rate is always 100\%, so tracking failures are absent from this dataset.

\textbf{ConsInv-Indoors-Dynamic subset.} \cref{tab:sota_all} shows that our approach outperforms all current methods. DynaSLAM and Slamantic perform worse than the Full masks baseline: thus, masking as much as possible (DynaSLAM does semantic + motion-based masking) or masking with prior geometric criteria (Slamantic) is suboptimal in difficult scenarios; masking appropriately is challenging. Lucas-Kanade optical flow has a low score as it does not avoid all consensus inversions and removes features during fast motions, lowering accuracy.

\textbf{ConsInv-Indoors-Static subset.} We also evaluate the ability to prevent false starts on the ConsInv-Indoors-Static subset. \cref{tab:sota_all} shows that most methods, including ours, prevent all false starts. The underperformance of Slamantic shows that motion-based geometric criteria are vulnerable to motion consensus inversions.

\textbf{ConsInv-Outdoors dataset.} \cref{tab:sota_all} shows that our multiclass self-supervised method outperforms the state of the art (USM = 0.88). Surprisingly, almost all other approaches have the same score (USM $\approx$ 0.80). This means that the masking strategy has to be learned from the environment, or performance may be suboptimal. 

\textbf{Ours vs Full Masks.} The high performance of Full Masks shows that it is a good baseline, better than most Dynamic SLAM algorithms in difficult situations. Still, it often has a lowered Tracking Rate (\ie, it loses tracking), which is unacceptable in critical applications. Our approach has both a high tracking rate and high accuracy.

\textbf{USM vs. ATE RMSE/TR.} Results on ConsInv show that ATE RMSE and TR alone do not reflect SLAM perf. meaningfully in difficult scenarios: \ie, low TR + low ATE RMSE and high TR + high ATE RMSE. The USM balances those meaningfully for all datasets. This also shows that ATE RMSE is not always more important than TR / USM.

\textbf{Qualitative results and performance.} The objects of a frame are masked when they are likely to move, \eg, when they were seen in the past of the same sequence. The masking inference can be done in real-time. Details are in supplementary materials.

\textbf{Global conclusion.} Our method outperforms the state of the art in all datasets. A remarkable result is that when object motion is difficult to detect as in the ConsInv dataset, the simple semantic approach \emph{Full Masks} is safer than motion-based, geometry-dependent approaches. Our method does not depend on geometric priors and outperforms \emph{Full Masks}, unlike current methods. And the USM makes comparisons more meaningful.

\subsection{Limitations: tests in out-of-context}
While we designed our method to adapt a SLAM to any context (same object classes/similar environment) at a low cost with self-supervision, we evaluated how the model trained on ConsInv-Indoors performs in the contexts of the ConsInv-Extra dataset -- a living room and a meeting room. \cref{tab:sota_all} shows that we perform slightly better than other SLAM algorithms. Thus, our model works in new contexts similar to the training one and with the same classes -- \ie, it supports domain shifts to a certain extent. If the change is more drastic, performance would likely decrease and require adaptations that are out of the scope of this paper. Although we mask all objects of the same class at once, with no instance separation, we still achieve state-of-the-art results in difficult scenarios, but performance may not be optimal if multiple objects of the same class cover the majority of an image with incoherent motions (\eg, a busy car parking).

\section{Conclusion}
In this paper, we introduced the paradigm of Temporal Masking to overcome the points of failures of current Dynamic SLAM algorithms. Given a source of semantic masks, the general idea is to apply semantic masks to filter objects of certain classes, when appropriate, to maximize a chosen SLAM metric. In detail, we proposed a memory-based neural network architecture that we train in a self-supervised way thanks to the proposed automatic annotation method. We compared the two obtained models with the state of the art \cite{bescos_dynaslam:_2018,schorghuber_slamantic_2019,cheng_improving_2019,scona_staticfusion:_2018} and outperformed it on real data: on the TUM RGB-D, KITTI and the proposed ConsInv dataset, that include challenging indoors/outdoors scenarios designed to test the limits of Dynamic SLAM algorithms. For this comparison, we designed a unified metric named USM that balances trajectory accuracy and robustness to failures due to excessive masking.

We are considering several improvements in future works. The proposed approach makes masking decisions at the class level, as a refinement, it seems valuable to explore a decision strategy that considers class instances separately. We are also considering developing a method merging semantic and Temporal Masking. A method that outputs semantic masks only when it improves a given metric would make our Temporal Masking method end-to-end with potentially superior semantic segmentation and masking decisions.

Finally, the Temporal Masking paradigm could be used outside SLAM applications, as it is a method to maximize the performance of an algorithm that takes as input a time series by triggering a specific action when appropriate. For instance, in path planning algorithms, it could prevent the exploration of probable dead ends based on the previous frames.

\textbf{Acknowledgment} this work relied on the use of the FactoryIA cluster, financially supported by the Ile-de-France Regional Council.

\clearpage

\bibliography{ms}
\end{document}